\documentclass[letterpaper]{article} 
\usepackage[submission]{aaai23}  
\usepackage{times}  
\usepackage{helvet}  
\usepackage{courier}  
\usepackage[hyphens]{url}  
\usepackage{graphicx} 
\usepackage{natbib}  
\usepackage{caption} 
\usepackage{algorithm}
\usepackage{algorithmic}

\usepackage{amsmath}
\usepackage{amssymb}
\usepackage{booktabs}

\usepackage{newfloat}
\usepackage{listings}
\DeclareCaptionStyle{ruled}{labelfont=normalfont,labelsep=colon,strut=off} 
\floatstyle{ruled}
\newfloat{listing}{tb}{lst}{}
\floatname{listing}{Listing}
\title{A Sequence Tagging based Framework for Few-Shot Relation Extraction}
\author {
    Xukun Luo and 
    Ping Wang
}
\affiliations {
    Peking University\\
    luoxukun@pku.edu.cn, pwang@pku.edu.cn
}

\usepackage{bibentry}

\begin{document}

\maketitle

\begin{abstract}
Relation Extraction (RE) refers to extracting the relation triples in the input text. Existing neural work based systems for RE rely heavily on manually labeled training data, but there are still a lot of domains where sufficient labeled data does not exist. Inspired by the distance-based few-shot named entity recognition methods, we put forward the definition of the few-shot RE task based on the sequence tagging joint extraction approaches, and propose a few-shot RE framework for the task. Besides, we apply two actual sequence tagging models to our framework (called Few-shot TPLinker and Few-shot BiTT), and achieves solid results on two few-shot RE tasks constructed from a public dataset.
\end{abstract}
\section{Introduction}

Relation Extraction (RE) is a fundamental task in natural language processing, aiming to extract the relation triples of the form $(h, r, t)$ in unstructured text. In a relation triple, $h$ and $t$ are the entities in the text, called the head entity and the tail entity respectively according to their positions. And $r$ describes the directed relationship from the head entity to the tail entity. Currently, the mainstream solutions follow the idea of joint extraction, which are mainly divided into the two methods: building up a label predicting system with a sequence tagging scheme \cite{Zheng2017Joint,Wei2020CasRel,Wang2020Tplinker} and constructing a seq2seq system \cite{Zeng2018Extracting,Nayak2019ptrnetdecoding,Zeng2020copymtl}. These systems have achieved solid results on the generic RE datasets with a sufficient amount of labeled instances.

However, one challenge remaining is that these RE systems based on the neural network require a lot of labeled training data, while the tagging effort is labor-intensive. The distant supervision method \cite{Mintz2009Distant} was proposed to obtain training data more cost-effectively. Based on existing knowledge graphs, the instances in the corresponding domain are automatically annotated. But a serious problem is that the samples labeled by the distant supervision method suffer from a large number of missed and incorrect tags. Thus the models must be reconstructed to accommodate these datasets with 
noise information. In addition, if there is no available knowledge graph for a particular domain, it is still impossible to complete the annotation for the training data.

As a result, researchers started to explore how to transfer knowledge from the resource-rich domains (such as the news domain) to the resource-poor domains (such as the medical domain). They worked to make the model achieve better results based on a small amount of labeled data in the target domain. Previously, they followed the simple pipeline idea and respectively assigned the knowledge transferred task to RE's two sub-tasks, i.e. named entity recognition (NER) \cite{Fritzler2019Few,Hou2020Few,Yang2020Simple} and relation classification (RC) \cite{Han2018FewRel,Gao2019RewRel,Baldini2019Matching}. These works are all distance-based approaches, drawing on the classical few-shot approaches in the image field. However, the error propagation of pipeline is self-explanatory. Currently, the methods of few-shot NER and few-shot RC are not effective enough, let alone combining the two for RE. 

Besides, the few-shot works in named entity recognition are essentially to find the closest label to a particular token of the input text. The idea is also applicable to the sequence label predicting systems in the joint extraction methods. Therefore, it is effective and feasible to apply the few-shot methods to joint extraction models.

In this paper, we propose a method\footnote{Available at \url{https://anonymous/for/review}.} to combine a sequence tagging based model for joint extraction and the few-shot methods to improve the performance with limited target domain's labeled data. In order to adopt the idea of distance-based few-shot sequence tagging to joint extraction methods, we begin with proposing a definition of the few-shot RE task through the sequence tagging based approaches. Second, we construct a sequence tagging based few-shot RE framework for the few-shot RE task. In our framework, the input instances are encoded by a BERT Encoder and an Adaptive Encoder, and then the distance between the hidden states of tokens and the labels are calculated, finally the predicted label sequence is outputted through distance comparison. Third, we apply two actual sequence tagging RE models \cite{Wang2020Tplinker,Luo2020BiTT} to the framework. These two models achieve solid results on two few-shot RE tasks constructed from a public dataset.

The key contributions are summarized as:
\begin{itemize}
    \item The definition of the few-shot RE task based on the sequence tagging approaches is presented for the first time.
    \item A few-shot joint RE framework is developed for sequence tagging approaches, and two actual sequence tagging models is applied to the framework.
    \item An accelerated approximation scheme is proposed for calculating the distance matrices in the training phrase of Section \ref{subsection:TPLinker}.
\end{itemize}

\section{Related Work}

\subsection{Relation Extraction}

RE, as a sub-task of information extraction, is an important part of building knowledge graphs. Current research on supervised methods is becoming increasingly sophisticated.

Some supervised RE methods followed the pipeline ideas \cite{Zeng2014Relation,Xu2015Classifying,Vu2016Combining,Zhong2021Frustratingly}, i.e., one model is used to recognize entities in a sentence, and then another model is used to identify the relations between each entity pair. The two models have different structures and parameters. Due to the error propagation problem of pipeline, other models turned to the joint method. They extracted entities and relations separately through two models that share parameters, or even simultaneously using a single model. For example, \cite{Miwa2016End,Katiyar2016Investigating} extended the relation classification module to share the encoder representation with the entity recognition module; \cite{Zheng2017Joint,Dai2019Joint,Wei2020CasRel,Luo2020BiTT,Wang2020Tplinker} took the extracting tasks as a sequence labeling problem, worked on proposing reliable tagging scheme that can reduce out relational triples; \cite{Zeng2018Extracting,Nayak2019ptrnetdecoding,Zeng2020copymtl} directly generated the triples by the seq2seq framework.

\subsection{Few-shot Relation Extraction}

The above supervised methods rely on a large amount of training data in the target domain. However, the labeling of relational triples in sentence is labor-intensive, resulting in frequent shortages of target domain's training data. Thus, learning from limited new labeled samples, called few-shot relation extraction, is a non-trivial task. To simplify the task, previous researchers followed the pipeline idea, which is different from the few-shot joint RE work in our paper.

\paragraph{Few-shot NER}
There are already a few approaches for Few-shot NER. An intuitive one is to extract cross-domain generic knowledge based on the source domain's labeled data, and then utilize target domain's limited data for knowledge migration. To carry on, \citet{Fritzler2019Few} applied the prototypical network \cite{Snell2017Prototypical} for Few-shot NER, utilizing prototypes comparison to tag tokens. \citet{Hou2020Few} put forward the collapsed dependency transfer mechanism, adapting the traditional conditional random field (CRF) to the mutable entity label categories. \citet{Yang2020Simple} proposed the nearest neighbor comparison scheme and pre-trained the few-shot NER model by a standard supervised NER method. \citet{Tong2021Learning} found and explored the rich semantics in $O$ class. And \citet{Ding2021Few} presented a practical and challenging dataset named Few-NERD for few-shot NER.

\paragraph{Few-shot RC}
Given one entity pair that appear in a sentence, the RC task becomes a multi-classification problem. \citet{Han2018FewRel} were the first to formalize the definition of few-shot RC and proposed a new dataset called FewRel. They apply several few-shot learning methods to FewRel, e.g., prototypical network and meta network \cite{Munkhdalai2017Meta}. \citet{Gao2019RewRel} built up FewRel2.0 for few-shot RC with two real-world issues, i.e., few-shot domain adaptation and few-shot none-of-the-above detection. \citet{Baldini2019Matching} employed a simple prototype approach for a CNN-based system. What's more, a more in-depth analysis \cite{Brody2021Towards} is performed on above works in a more realistic RE setting.
\section{Problem Definition}

\begin{algorithm}[t]
    \caption{$N$-way $K\sim2K$-shot Sampling}
    \label{alg:sample}
    \begin{algorithmic}[1]
        \REQUIRE
            Dataset $\mathbf{X}$;
            Relation Category set $\mathcal{C}$;
            $N$ and $K$;
        \ENSURE
            A support set $\mathcal{S}$
        \STATE Initialize $\mathcal{S} \gets \Phi$;
        \FOR {each $i \in \{1, ..., N\}$}
            \STATE $Count_i \gets 0$;
        \ENDFOR
        \WHILE {$\exists Count_i \le K, 1 \le i \le N$}
            \STATE Randomly sample $(x, \mathcal{R}) \in \mathbf{X}$;
            \STATE Calculate $Count_i$ after update;
            \IF {$Count_i \le 2K, \forall i \in \{1, ..., N\}$}
                \FOR {each $i \in \{1, ..., N\}$}
                    \STATE $\boldsymbol{y}_i \gets \mathcal{T}(\boldsymbol{x}, \mathcal{R}, i)$;
                \ENDFOR
                \STATE $\mathcal{S} \gets \mathcal{S} \cup \{(\boldsymbol{x}, \boldsymbol{y}_1, ..., \boldsymbol{y}_N)\}$;
            \ENDIF
        \ENDWHILE
    \end{algorithmic}
\end{algorithm}

In the sequence tagging based approaches of joint extraction, the label sequence $\boldsymbol{y} = (y_1, y_2, ..., y_n)$ can be obtained by a specific invertible tagging function $\mathcal{T}(\boldsymbol{x}, \mathcal{R}, r)$, where $\boldsymbol{x} = (x_1, x_2, ..., x_m)$ is a sequence of tokens. $\mathcal{R}$ indicates the set of relational triples in $\boldsymbol{x}$ and $r$ is a specific relation category. Note that the numerical relationship between $m$ and $n$ is related to function $\mathcal{T}$ from previous works, e.g., $n = m$ in \cite{Zheng2017Joint} and $n = m^2$ in \cite{Dai2019Joint}.

\begin{figure*}[t]
  \centering
  \includegraphics[width=.8\textwidth]{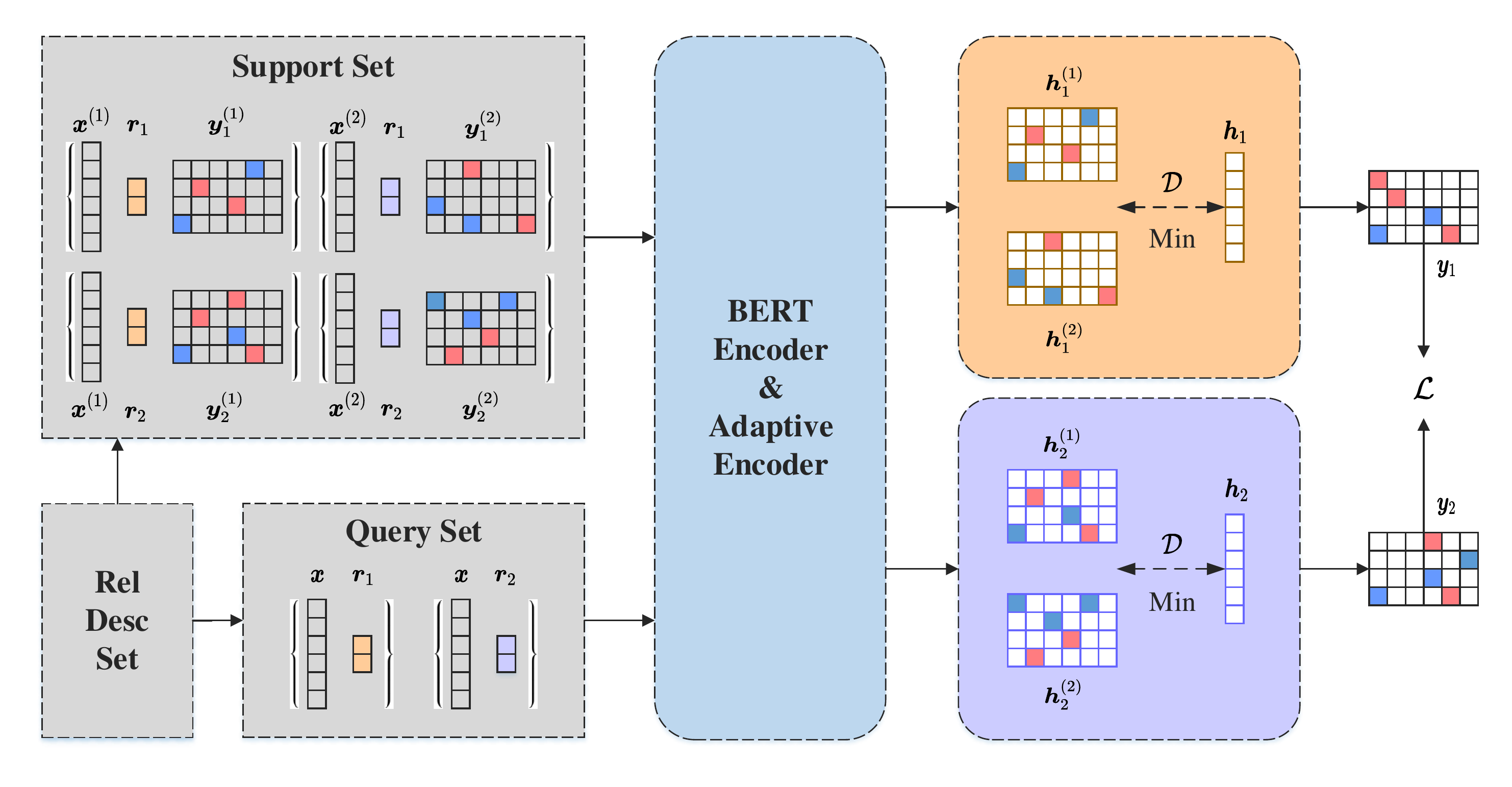}\\
  \caption{The sequence tagging based framework for few-shot RE. The colors in $\boldsymbol{y}$ correspond to different labels.}
  \label{Fig:Framework}
\end{figure*}

Thus in few-shot RE, we intent to obtain a function $\mathcal{F} : (\mathcal{C}, \mathcal{S}, \mathcal{Q}) \mapsto \{\boldsymbol{y}_1, ..., \boldsymbol{y}_{|\mathcal{C}|}\}$. Here $\mathcal{C}$ indicates a set of $N$ relation categories. And $\mathcal{S} = \{(\boldsymbol{x}^{(i)}, \boldsymbol{y}_1^{(i)}, ..., \boldsymbol{y}_{N}^{(i)})\}_{i=1}^{N_s}$ is called the support set, including $N$ relation categories ($N$-way) and $K$ instances ($K$-shot) for each category in the target domain. However, there may be multiple categories of triples in a single instance, making it difficult to sample from the source domain when training. To solve the problem, similar to \cite{Ding2021Few}, we introduce the $N$-way $K\sim2K$-shot setting into few-shot RE task (Algorithm \ref{alg:sample}), i.e., a sampled $\mathcal{S}$ contains $K\sim2K$ instances for each relation category. Besides, $\mathcal{Q}$ is called the query set, which is the collection of the token sequences that need to be heuristically labeled according to $\mathcal{S}$. For simplicity, the number of instances in $\mathcal{Q}$ is assumed to be 1, i.e., $|\mathcal{Q}| = 1$. And $\{\boldsymbol{y}_1, ..., \boldsymbol{y}_{|\mathcal{C}|}\}$ is the prediction of the single instance $\boldsymbol{x}$ in $\mathcal{Q}$.

\section{Methodology}
\label{section:method}

In this section, we first present our sequence tagging based few-shot RE framework in general. Then we give two examples of how to apply the previous sequence tagging models for joint extraction to our framework, i.e., Few-shot TPLinker \cite{Wang2020Tplinker} and Few-shot BiTT \cite{Luo2020BiTT}.

\subsection{Few-shot Relation Extraction Framework}

As shown in Figure \ref{Fig:Framework}, our few-shot RE framework for sequence tagging approaches consists of the following 4 steps. 

First, for each relation category, our framework needs to generate a label sequence by function $\mathcal{T}$ for each instance in $\mathcal{S}$. For example, $\boldsymbol{y}_1^{(1)}$ in Figure \ref{Fig:Framework} is the label sequence derived from the instance $\boldsymbol{x}^{(1)}$ in $\mathcal{S}$ and the relation triple set in $\boldsymbol{x}^{(1)}$ with relation category $\boldsymbol{r_1}$.

Second, each token sequence $\boldsymbol{x}$ in $\mathcal{S}$ and $\mathcal{Q}$ along with the description of a specific relation $\boldsymbol{r}$ are input into a BERT Encoder \cite{Devlin2019BERT} to get the embedding of each token in $\boldsymbol{x}$, as follow. 
\begin{align}\label{equa:Embedding}
  \mathbf{E} = BertEncoder([\boldsymbol{r}; \boldsymbol{x}])
\end{align}
Note that in Eq.\eqref{equa:Embedding} we utilize \texttt{[SEP]} as the concatenation token of $\boldsymbol{r}$ and $\boldsymbol{x}$.

Third, for different sequence tagging models, our framework sets up an Adaptive Encoder, allowing models to obtain the hidden states for each position to be labeled that meet their own requirements.
\begin{align}\label{equa:AdaptiveEncoder}
  \mathbf{H} = AdaptiveEncoder(\mathbf{E})
\end{align}
Here the dimension of $\mathbf{H}$ is $n \times N_h$, where $n$ is related to the tagging strategy $\mathcal{T}$ and $N_h$ is the pre-defined hidden size. Note that the parameters of the Bert Encoder and the Adaptive Encoder can be pre-trained in advance with a traditional sequence tagging task in the source domain by appending a linear layer after $\mathbf{H}$ and maximizing the probability of every correct label.

Finally, we compute the distance of the hidden state between each position of $\boldsymbol{y}$ in $\mathcal{S}$ and $\mathcal{Q}$. 
\begin{align}\label{equa:Distance}
  \mathbf{D} = \mathcal{D}(\mathbf{H}^{(\mathcal{Q})}, \mathbf{H}^{(\mathcal{S})})
\end{align}
Here $\mathbf{D} \in \mathbb{R}^{n \times n|\mathcal{S}|}$ is the distance matrix, where $|\mathcal{S}|$ means the instance number in $\mathcal{S}$. $\mathbf{H}^{(\mathcal{Q})}$ and $\mathbf{H}^{(\mathcal{S})}$ are the hidden states of $\mathcal{Q}$ and $\mathcal{S}$ respectively from Eq.\eqref{equa:AdaptiveEncoder}. Besides, $\mathcal{D}(\cdot)$ indicates the squared Euclidean distance function. Note that since $\boldsymbol{y}$ is chunked in some specific tagging strategies, the distances between positions across chunks are not calculated to avoid the confusion of the labels in different chunks. For example, in TPLinker \cite{Wang2020Tplinker}, there are three tagging matrices which contain different tags and represent different information. The labels sequence in a single matrix for all tokens is called a chunk and $\boldsymbol{y}$ generated based on the TPLinker tagging strategy contains three chunks.
For the inference stage, we utilize the label of the closest position $p_s$ in $\mathcal{S}$ to a specific position $p_q$ in $\mathcal{Q}$ as $p_q$'s label, as follow. 
\begin{align}\label{equa:Inference}
  &\boldsymbol{y}^*(p_q) = \boldsymbol{t}(p_s) \\
  &p_s = arg \min_{1 \le i \le n|\mathcal{S}|}{\mathbf{D}[p_q, i]}
\end{align}
Here $\boldsymbol{y}^*$ is the predicted label sequence of $\mathcal{Q}$ for the specific relation $\boldsymbol{r}$ and $\boldsymbol{t}(\cdot)$ indicates the mapping function from each position in $\mathcal{S}$ to its label.
And for the training stage, we assume that $\boldsymbol{y}$ is divided into $\lambda$ chunks of equal length. In a specific chunk with index $c \in \{0,\cdots,\lambda-1\}$, the distance between a position $p$ in $\mathcal{Q}$ and a label $l$ can be calculated by:
\begin{align}\label{equa:Train}
  \mathbf{D}_c^l[p] = \min_{p^* \in \mathcal{W}_c^l}{\mathbf{D}[p, p*]}
\end{align}
where $\mathcal{W}_c^l$ indicates the set of all positions labeled with $l$ in the chunk $c$ of $\mathcal{S}$. Note that the chunk $c$ contains $N_c$ kinds of labels, i.e., $l \in \{0,\cdots,N_c-1\}$. And we apply the cross entropy loss function for every chunk to minimize the distances between $p$ and the correct label, as follow.
\begin{align}\label{equa:LossChunk}
  &\mathcal{L}_c = - \frac{\lambda}{n{N_c}}\sum_{i=1}^{n/{\lambda}}\sum_{l=1}^{N_c}(\hat{\boldsymbol{y}}_{\frac{c{n}}{\lambda}+i}(l)\mathcal{P}(\mathbf{D}_c^l[i])) \\
  &\mathcal{P} (\mathbf{D}_c^l[i]) = log(\frac{e^{-\mathbf{D}_c^l[i]}}{\sum_{l^\prime=1}^{N_c}e^{-\mathbf{D}_c^{l^\prime}[i]}})
\end{align}
Here $\hat{\boldsymbol{y}}_{\frac{c{n}}{\lambda}+i} \in \mathbb{R}^{N_c}$ is the one-hot form of $y_{\frac{c{n}}{\lambda}+i}$, which means the label index of the $i$-th position in chunk $c$ of $\boldsymbol{y}$. The final loss $\mathcal{L}$ is the average of $\mathcal{L}_c$.


\subsection{Few-shot TPLinker}
\label{subsection:TPLinker}

\begin{figure}[t]
  \centering
  \includegraphics[width=\columnwidth]{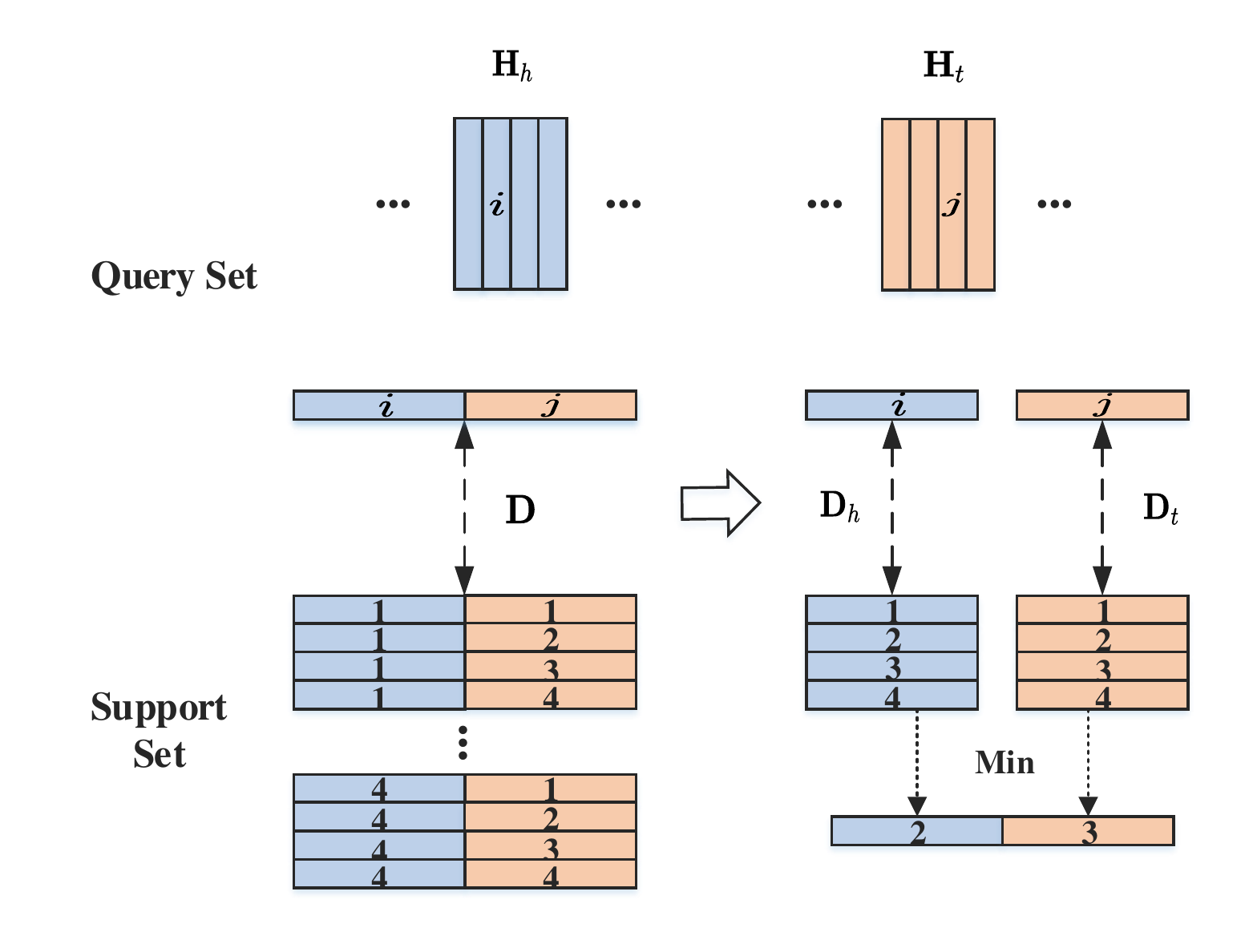}\\
  \caption{An example of calculating the distances between ($x_{i}$, $x_{j}$) in $\mathcal{Q}$ and the token pairs in $\mathcal{S}$ for Few-shot TPLinker. Assume that ($x_{i}$, $x_{j}$) in $\mathcal{Q}$ and ($x_2$, $x_3$) in $\mathcal{S}$ have the smallest distance. And the number of tokens in $\mathcal{S}$ is 4.}
  \label{Fig:FewTPLinker}
\end{figure}

TPLinker \cite{Wang2020Tplinker} is an one-stage joint extraction model. It predicts three tagging matrices for the input instance, i.e., EH-to-ET, SH-to-OH and ST-to-OT. The elements in the matrices are selected in $\{0, 1, 2\}$. Note that the label $2$ is proposed to reduce the matrix space by mapping the label $1$ in the lower triangular region to the upper triangular region. However, in the few-shot RE task, there may be no label $2$ in the tagging matrices of $\mathcal{S}$, making it hard to calculate the distance from the positions to be annotated to label $2$. Therefore, we cancel this improvement when applying TPLinker to our framework. In this case, the length of the label sequence $\boldsymbol{y}$ for a specific relation category is $3m^2$, and $\boldsymbol{y}_i \in \{0, 1\}$ where $ 1 \le i \le 3m^2$. And the number of chunks in TPLinker is $\lambda = 3$. The following describes the process for a single chunk $c$ of applying TPLinker to the sequence tagging based few-shot RE framework.

\paragraph{Adaptive Encoder}
For the optimization of the next distance calculation step, we define the hidden state $\mathbf{H}[i*m + j]$ of ($x_i$, $x_j$) as a concatenation of two parts, as follow.
\begin{align}\label{equa:TPLinkerHidden}
  \mathbf{H}[i{m} + j] &= [\mathbf{H}_h[i]; \mathbf{H}_t[j]]
\end{align}
Here $\mathbf{H}_h[i]$ and $\mathbf{H}_t[j]$ respectively denote the hidden states of the head token $x_i$ and the tail token $x_j$. They are calculated by two linear layers:
\begin{align}\label{equa:TPLinkerHiddenH}
  \mathbf{H}_h &= \mathbf{W}_h\mathbf{E}[i] + \boldsymbol{b}_h \\\label{equa:TPLinkerHiddenT}
  \mathbf{H}_t &= \mathbf{W}_t\mathbf{E}[j] + \boldsymbol{b}_t
\end{align}

\begin{figure}[t]
  \centering
  \includegraphics[width=\columnwidth]{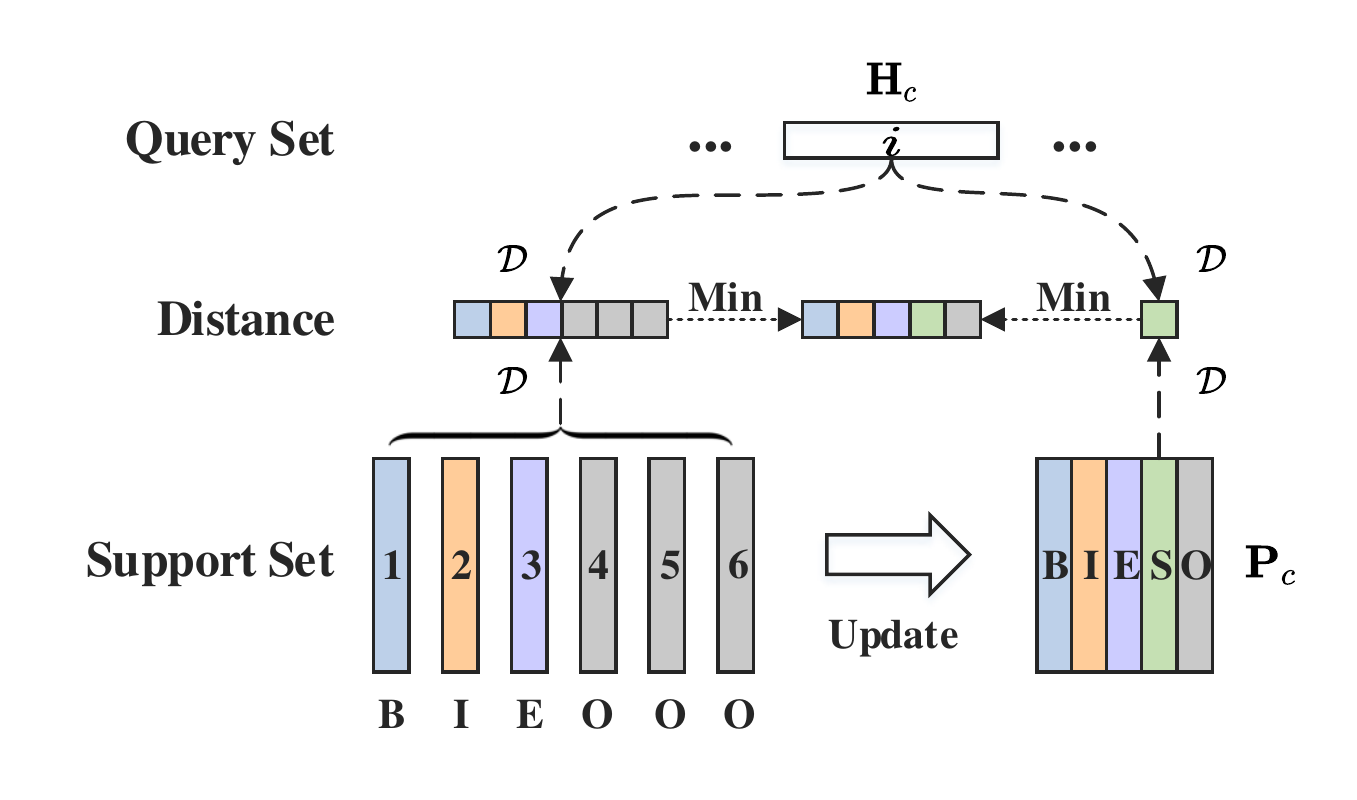}\\
  \caption{An example of calculating the distances between $x_{i}$ in $\mathcal{Q}$ and the tokens in $\mathcal{S}$ for Few-shot BiTT. The chunk $c$ is the first BiTT part which labels the entities through `BIESO'. The hidden vectors in the same color belong to one tag. The operations Min and Update are separately performed among the distances and vectors with the same color.}
  \label{Fig:FewBiTT}
\end{figure}

\begin{table*}[t]
\centering
\resizebox{\textwidth}{!}{
\begin{tabular}{@{}clclc@{}}
\toprule
\textbf{Group} &  & \textbf{Categories} &  & \textbf{Instances Num} \\ \midrule
A              &  & \textit{\begin{tabular}[c]{@{}c@{}}/people/person/ethnicity, /location/location/contains, /sports/sports\_team\_location/teams, \\ /business/company/founders, /people/person/nationality, /business/company/advisors, \\ /business/person/company, /location/country/capital\end{tabular}}                                                           &  & 56,546                  \\ \midrule
B              &  & \textit{\begin{tabular}[c]{@{}c@{}}/people/person/place\_lived, /business/company\_shareholder/major\_shareholder\_of, /people/ethnicity/people, \\ /location/neighborhood/neighborhood\_of, /business/company/major\_shareholders, /people/person/place\_of\_birth, \\ /business/company/place\_founded, /sports/sports\_team/location\end{tabular}} &  & 14,942                  \\ \midrule
C              &  & \textit{\begin{tabular}[c]{@{}c@{}}/location/administrative\_division/country, /location/country/administrative\_divisions, /people/person/profession, \\ /people/ethnicity/geographic\_distribution, /people/person/religion, /people/person/children, \\ /business/company/industry, /people/deceased\_person/place\_of\_death\end{tabular}}        &  & 8,284                   \\ \bottomrule
\end{tabular}}
\caption{Statistics of the groups in Few-NYT (INTER).}\label{Tab:Statistics}
\end{table*}

\paragraph{Distance Matrix}
As shown in Figure \ref{Fig:FewTPLinker}, the distance between the token pairs ($x_{i}$, $x_{j}$) in $\mathcal{Q}$ and ($x_{i_s}$, $x_{j_s}$) in $\mathcal{S}$ is divided into two part, as follow.
\begin{align}\label{equa:TPLinkerDistance}
  \mathbf{D}[i{m}+j, i_s{m}+j_s] = \mathbf{D}_h[i, i_s] + \mathbf{D}_t[j, j_s]
\end{align}
Here $\mathbf{D}_h, \mathbf{D}_t \in \mathbb{R}^{m \times m|\mathcal{S}|}$ are distance matrices between the tokens in $\mathcal{Q}$ and $\mathcal{S}$. $\mathbf{D}_h$ and $\mathbf{D}_t$ are generated from $\mathbf{H}_h$ and $\mathbf{H}_t$ respectively by Eq.\eqref{equa:Distance}. Note that $x_{i_s}$ and $x_{j_s}$ belong to the same instance in $\mathcal{S}$.
In Few-shot TPLinker, the distances from ($x_i$, $x_j$) to label $1$ and label $0$, called $\mathbf{D}_p[i,j]$ and $\mathbf{D}_n[i,j]$, is required to be obtained respectively. If the solution is preformed by traversing all token pairs in $\mathcal{S}$, the overall time complexity for working out $\mathbf{D}_p$ and $\mathbf{D}_n$ can be $O(m^4|\mathcal{S}|)$. For example, as shown in the left part of Figure \ref{Fig:FewTPLinker}, all distances between ($x_i$, $x_j$) and 
all token pairs in $\mathcal{S}$ should be computed once, for a total of 16 times. Thus, we devise an accelerated approximation scheme in the training phrase. First, iterate over the positive samples set $\mathcal{W}_\mathcal{S}$ to fill $\mathbf{D}_p$ as follow, since $|\mathcal{W}_\mathcal{S}| \ll |\mathcal{S}|$.
\begin{equation}\label{equa:TPLinkerTrainTagPos}
  \begin{aligned}
    \mathbf{D}_p[i, j] = \min_{(x_{i_s}, x_{j_s}) \in \mathcal{W}_\mathcal{S}}({\mathbf{D}_h[i, i_s] + \mathbf{D}_t[j, j_s]})
  \end{aligned}
\end{equation}
Second, assume that there is at least one negative sample among the top-$\mathcal{E}$ tokens pairs with the shortest distance from ($x_i$, $x_j$). These top-$\mathcal{E}$ distances $\hat{\mathbf{D}}[i,j] \in \mathbb{R}^{\mathcal{E}}$ can be given by:
\begin{equation}\label{equa:TPLinkerTrainTopK1}
  \begin{aligned}
    \hat{\mathbf{D}}[i,j] = top_\mathcal{E}(\hat{\boldsymbol{a}}_i^s(\varepsilon_1) + \hat{\boldsymbol{b}}_j^s(\varepsilon_2))
  \end{aligned}
\end{equation}
where $1 \le s \le |\mathcal{S}|$, $1 \le \varepsilon_1, \varepsilon_2 \le \mathcal{E}$, and $\mathcal{E} \ge 2$. $\hat{\boldsymbol{a}}_i^s \in \mathbb{R}^{\mathcal{E}}$ and $\hat{\boldsymbol{b}}_j^s \in \mathbb{R}^{\mathcal{E}}$ denote the distances respectively from $x_i$'s and $x_j$'s top-$\mathcal{E}$ nearest tokens of the s-th instance in $\mathcal{S}$.
\begin{align}\label{equa:TPLinkerTrainTopK2}
  \hat{\boldsymbol{a}}_i^s &= top_\mathcal{E}(\mathbf{D}_h[i, s{m}+a]), 1 \le a \le m \\\label{equa:TPLinkerTrainTopK3}
  \hat{\boldsymbol{b}}_j^s &= top_\mathcal{E}(\mathbf{D}_t[j, s{m}+b]), 1 \le b \le m
\end{align}
Last, denote the set of values in $\hat{\mathbf{D}}[i,j]$ except for $\mathbf{D}_p[i,j]$ as $\mathcal{W}$. $\mathbf{D}_n$ can be roughly filled with the minimum (Eq.\eqref{equa:TPLinkerTrainTagNegTop}) or average (Eq.\eqref{equa:TPLinkerTrainTagNegAvg}) value in $\mathcal{W}$. Thus the time complexity of our accelerated approximation scheme for $\mathbf{D}_p$ and $\mathbf{D}_n$ is $O(m^2(|\mathcal{W}_\mathcal{S}| + \mathcal{E}))$.
\begin{align}\label{equa:TPLinkerTrainTagNegTop}
  \mathbf{D}_n[i, j] &= \left\{\begin{matrix}
    \hat{\mathbf{D}}[i, j, 1], \mathbf{D}_p[i, j] \neq \hat{\mathbf{D}}[i, j, 1] \\
    \hat{\mathbf{D}}[i, j, 2], \mathbf{D}_p[i, j] = \hat{\mathbf{D}}[i, j, 1]
  \end{matrix}\right. \\\label{equa:TPLinkerTrainTagNegAvg}
  \mathbf{D}_n[i, j] &= \frac{\sum_{\hat{\mathbf{D}}[i, j, \varepsilon] \in \mathcal{W}}\hat{\mathbf{D}}[i, j, \varepsilon]}{|\mathcal{W}|}
\end{align}

\subsection{Few-shot BiTT}
\label{subsection:BiTT}

BiTT \cite{Luo2020BiTT} is an end-to-end RE framework with the bidirectional tree tagging scheme. It predicts a label sequence containing the tree structures of the relational triples for an input sentence, and then reconstructs the relational graph to extract triples. In BiTT scheme, the chunk number $\lambda$ is 8, and the length of $\boldsymbol{y}$ is $\lambda{m}$. The following introduces the procedure in a single chunk $c$ for adopting BiTT into our framework.

\paragraph{Adaptive Encoder}
We define that the hidden state of $y_{c*m+i}$ is related to the embedding of $x_i$ by the following equation:
\begin{align}\label{equa:BiTTHidden}
  \mathbf{H}_c[i] &= \mathbf{W}_c\mathbf{E}[i] + \boldsymbol{b}_c
\end{align}
where $\mathbf{W}_c$ is the weighted matrix and $\boldsymbol{b}_c$ is the bias. 

\paragraph{Distance Matrix}
As shown in Figure \ref{Fig:FewBiTT}, we calculate the distances from $x_i$ to the tokens in $\mathcal{S}$ and work out the minimum distance for each label. Since there may not be any token with label $l$, we calculate the distance between $x_i$ and a prototype $\mathbf{P}_c[l]$ inherited from previous training steps.
\begin{align}\label{equa:BiTTUpdate}
  \mathbf{P}_c[l] = (1 - \gamma)\mathbf{P}_c[l] + \gamma \frac{\sum_{x_i\in \mathcal{W}_c^l}{\mathbf{H}_c^{(\mathcal{S})}[i]}}{|\mathcal{W}_c^l|}
\end{align}
Here $\mathcal{W}_c^l$ indicates the set of all tokens labeled with $l$ in the chunk $c$ of $\mathcal{S}$. Note that the update operation is executed only if $\mathcal{W}_c^l \neq \Phi$.

\section{Experiments}

\subsection{Dataset and Evaluation Tasks}
Considering that different RE datasets are generated with different methods and relation category naming standards, it may face the issues of inconsistent relation descriptions and contradictory common knowledge when using multiple datasets for few-shot experiments. Moreover, it is practical to utilize a single dataset to perform different few-shot task scenarios, and then effectively validate the performance of our framework. Therefore, we evaluate Few-shot TPLinker and Few-shot BiTT on NYT \cite{Zeng2014Relation}, an English RE dataset with 66,195 instances. NYT is labeled by the distant supervision method and a hierarchical naming standard. It pre-defines 24 fine-grained relation categories and 4 coarse-grained relation types.

Following the few-shot task setting of \cite{Ding2021Few}, we split the overall relation triple set into groups on consideration for the different granularity of category names. In this case, every group corresponds to a sub-dataset of instances containing the triples in the group. Besides, to avoid the observation of triples belonging to other groups, we need to mask out these triples first and then perform sequence tagging on the instances. Based on this setting, we develop two few-shot RE tasks adopting different splitting scheme and conduct 2-way 1$\sim$2-shot experiments on these tasks.

\begin{table*}[t]
\centering
\resizebox{\textwidth}{!}{
\begin{tabular}{ccccccccccccccccccccccccc}
\hline
\textbf{Model}                    &  & \multicolumn{5}{c}{\textbf{Group A}}                &  & \multicolumn{5}{c}{\textbf{Group B}}                &  & \multicolumn{5}{c}{\textbf{Group C}}                &  & \multicolumn{5}{c}{\textbf{Average}}                \\ \cline{3-7} \cline{9-13} \cline{15-19} \cline{21-25} 
                                  &  & \textit{Prec} &  & \textit{Rec}  &  & \textit{F1}   &  & \textit{Prec} &  & \textit{Rec}  &  & \textit{F1}   &  & \textit{Prec} &  & \textit{Rec}  &  & \textit{F1}   &  & \textit{Prec} &  & \textit{Rec}  &  & \textit{F1}   \\ \hline
\textit{No pretrain, no finetune} &  &               &  &               &  &               &  &               &  &               &  &               &  &               &  &               &  &               &  &               &  &               &  &               \\
Few-shot TPLinker (min)           &  & 24.2          &  & 15.1 &  & 18.6          &  & 44.8          &  & 26.1 &  & 33            &  & 22.5          &  & 22.9 &  & 22.7 &  & 30.5          &  & 21.4 &  & 24.8 \\
Few-shot TPLinker (avg)           &  & 6.4           &  & 13.1          &  & 8.6           &  & 28.9          &  & 22.9          &  & 25.6          &  & 4             &  & 18            &  & 6.7           &  & 13.1          &  & 18            &  & 13.6          \\
Few-shot BiTT                     &  & 47.7 &  & 12.9          &  & 20.4 &  & 51.6 &  & 14.4          &  & 22.5          &  & 23.6 &  & 8.3           &  & 12.3          &  & 41   &  & 11.9          &  & 18.4          \\ \hline
\textit{Pretrain, no finetune}    &  &               &  &               &  &               &  &               &  &               &  &               &  &               &  &               &  &               &  &               &  &               &  &               \\
Few-shot TPLinker (min)           &  & 38.7          &  & 12.5 &  & 18.9 &  & 50            &  & 22.7          &  & 31.2          &  & 33.4          &  & 14.3          &  & 20            &  & 40.7          &  & 16.5          &  & 23.4          \\
Few-shot TPLinker (avg)           &  & 37.8          &  & 12.3          &  & 18.5          &  & 46.9          &  & 22.4          &  & 30.3          &  & 25.9          &  & 20   &  & 22.6 &  & 36.9          &  & 18.2 &  & 23.8 \\
Few-shot BiTT                     &  & \textbf{63.9} &  & 10.5          &  & 18.1          &  & \textbf{64}   &  & 22.8 &  & 33.6 &  & 53.4 &  & 6.4           &  & 11.5          &  & \textbf{60.4} &  & 13.2          &  & 21.1          \\ \hline
\textit{Pretrain, finetune}       &  &               &  &               &  &               &  &               &  &               &  &               &  &               &  &               &  &               &  &               &  &               &  &               \\
Few-shot TPLinker (min)           &  & 21.3          &  & \textbf{18.7} &  & 19.9          &  & 50.1          &  & \textbf{33.6} &  & \textbf{40.2} &  & 37.3          &  & \textbf{24}   &  & \textbf{29.2} &  & 36.2          &  & \textbf{25.4} &  & \textbf{29.8} \\
Few-shot TPLinker (avg)           &  & 43            &  & 17            &  & \textbf{24.3} &  & 48.5          &  & 31.5          &  & 38.2          &  & 25.9          &  & 20            &  & 22.6          &  & 39.1          &  & 22.8          &  & 28.4          \\
Few-shot BiTT                     &  & 60.5 &  & 13.4          &  & 21.9          &  & 62.2 &  & 27.2          &  & 37.9          &  & \textbf{54.7} &  & 10.3          &  & 17.3          &  & 59.1 &  & 17            &  & 25.7          \\ \hline
\end{tabular}}
\caption{Main results on the Few-NYT (INTER) task. The best results are in \textbf{bold}.}\label{Tab:InterResult}
\end{table*}

\paragraph{Few-NYT (INTER)}
In this task, we randomly and manually divide the 24 relation categories in NYT into group A, group B and group C. As shown in Table \ref{Tab:Statistics}, every group is assigned 8 fine-grained relation categories, and the instances number of three sub-datasets are 56,546, 14,942 and 8,284 respectively. We do not consider the coarse-grained type to which each fine-grained relation category belongs. For example, \textit{/people/person/ethnicity}, \textit{/people/person/place\_lived} and \textit{/people/person/children} all belong to the coarse-grained type \textit{people}, while they are separately partitioned into group A, group B and group C. In the experiments of this task, we take each group as the target domain for evaluating our models, and the remaining two groups as the source domain for training. In Few-NYT (INTER), our models may be more capable of mining the common knowledge between the source and target domains, since the three groups share some of the coarse information.

\paragraph{Few-NYT (INTRA)}
In NYT, there are 4 predefined coarse-grained relation types, i.e., \textit{location}, \textit{business}, \textit{sports} and \textit{people}. When dividing these four relation types, we consider about the balance the amount of data in the training and validation sets, and also expect to mask out as few as possible relation triples. Thus, we simply adopt the instances set of \textit{location}, \textit{business} and \textit{sports} as the training set, and the instances set of \textit{people} as the validation set. The instances number of the training and validation sets are 51,559 and 16,711 respectively. In contrast to Few-NYT (INTER), the source and target domains of Few-NYT (INTRA) do not share coarse information as there is no intersection of coarse-grained relation types. Therefore, our models may be able to explore less inter-domain pervasive information, making this task more difficult.

\subsection{Experimental Settings}
\paragraph{Competitive Models}
To the best of our knowledge, our Few-shot Relation Extraction Framework is the first few-shot system for joint extraction of entities and relations. Thus we do not choose other baselines, but construct three competitive models based on our framework for comparison experiments, as shown in Tabel \ref{Tab:InterResult} and Tabel \ref{Tab:IntraResult}.
\begin{itemize}
    \item \textbf{Few-shot TPLinker (min)}: The Few-shot TPLinker model mentioned in Section \ref{subsection:TPLinker}, which applys Eq.\eqref{equa:TPLinkerTrainTagNegTop} to the calculation of the negative distance matrix $\mathbf{D}_n$.
    \item \textbf{Few-shot TPLinker (avg)}: The Few-shot TPLinker model mentioned in Section \ref{subsection:TPLinker}, applying Eq.\eqref{equa:TPLinkerTrainTagNegAvg} to the calculation of $\mathbf{D}_n$.
    \item \textbf{Few-shot BiTT}:  The Few-shot BiTT model mentioned in Section \ref{subsection:BiTT}.
\end{itemize}
In addition, we design three training modes for these models to explore the effect of pre-training and fine-tuning on them. \textit{No pretrain, no finetune} indicates that the models are trained directly from scratch based on the loss (Eq.\eqref{equa:LossChunk}) of our few-shot RE framework. \textit{Pretrain, no finetune} indicates that the models are only pre-trained in the traditional sequence tagging RE task, while not fine-tuned in the few-shot RE task. \textit{Pretrain, finetune} indicates that the models are both pre-trained in the traditional sequence tagging RE task and fine-tuned in the few-shot RE task.

\paragraph{Parameter Tuning}
In our framework, we adopt the default BERT-Based-Cased hyper-parameter values provided by Hugging Face\footnote{\url{https://huggingface.co/}}. Overall, in our experiments, the number of iterations for direct training and fine-tuning are 500,000 and 100,000 separately, and the number of epochs in pre-training is 20. In the process of direct training and fine-tuning, 5,000 iterations of validation are performed for every 10,000 iterations of training. In the pre-training process, we divide the data in the source domain into training set and validation set in the ratio of 4:1, and evaluate on the validation set after two training epochs. Then we save the pre-trained models with the best result. Besides, the batch size in pre-training process is set to 128, and the learning rate of Adam optimizer is 2e-5.

What's more, there are a few hyper-parameters that need to be set artificially in the formulations elaborating our framework and models in Section \ref{section:method}. For our few-shot RE framework, in Eq.\eqref{equa:Embedding}, the maximum length of the relation category description and the input tokens sequence are 10 (including \texttt{[CLS]} and \texttt{[SEP]}) and 50 (including \texttt{[SEP]}) respectively. And the relation category description $\boldsymbol{r}$ is the word sequence obtained from the corresponding fine-grained relation category name by the tokenization based on the BERT-Based-Cased vocabulary. In Eq.\eqref{equa:AdaptiveEncoder}, the hidden size of $\mathbf{H}$ is $N_h = 32$. For few-shot TPLinker, in Eq.\eqref{equa:TPLinkerHiddenH} and Eq.\eqref{equa:TPLinkerHiddenT}, the dimension of $\mathbf{W}_h$ and $\mathbf{W}_t$ is 768$\times$32. And $\mathcal{E}$ is set to 3 in Eq.\eqref{equa:TPLinkerTrainTopK1}, Eq.\eqref{equa:TPLinkerTrainTopK2} and Eq.\eqref{equa:TPLinkerTrainTopK3}. For few-shot BiTT, in Eq.\eqref{equa:BiTTHidden}, the dimension of $\mathbf{W}_c$ is 768$\times$32. In Eq.\eqref{equa:BiTTUpdate}, the hidden size $\mathbf{P}_c$ is 32 and the weighted variable $\gamma$ is set to 0.9.

\paragraph{Evaluation Metrics}

The optimal models after training are evaluated on the target domain by 20,000 iterations of sampling. We report the standard evaluation metrics for RE: Precision (\textit{Prec}), Recall (\textit{Rec}) and F1 score (\textit{F1}). We consider a predicted triple $(h, r, t)$ as a correct one only if $h$, $t$ and $r$ are all correct.

\subsection{Results}
The experimental results on Few-NYT (INTER) and Few-NYT (INTRA) are shown in Table \ref{Tab:InterResult} and Table \ref{Tab:IntraResult} respectively.

\paragraph{Models Comparison}
As shown in Table \ref{Tab:InterResult}, each of the three models based on our proposed Few-shot RE framework has its own benefits. First, Few-shot TPLinker (min) achieves the best average \textit{F1} (29.8\%) and \textit{rec} (25.4\%) scores on Few-NYT (INTER), outperforms the second best model (Few-shot TPLinker (avg)) by 1.4\% and 2.6\% separately. This indicates that Few-shot TPLinker (min) works best overall and is able to recognize more valid relation triples in the target domain instances. Second, Few-shot TPLinker (avg) obtains the best \textit{F1} (24.3\%) scores on Few-NYT (INTER) when we take group A as the target domain. It outperforms Few-shot TPLinker (min) by 4.4\% and Few-shot BiTT by 2.4\% in \textit{F1} respectively when the amount of instances (23,226) in the source domain is comparatively small. Third, Few-shot BiTT achieves the best average \textit{Prec} (60.4\%) score on Few-NYT (INTER), outperforms the second model (Few-shot TPLinker (min)) by 19.7\%. This demonstrates that Few-shot BiTT tends to focus on the validity of its own predictions, rather than locating as many triples in the instance as possible. It is due to the fact that Few-shot BiTT only outputs a predicted triple when all tokens of both head entity and tail entity are consistently labeled.

\begin{table}[t]
\centering
\resizebox{\columnwidth}{!}{
\begin{tabular}{ccccccc}
\hline
\textbf{Model}                    &  & \textit{Prec} &  & \textit{Rec} &  & \textit{F1}  \\ \hline
\textit{No pretrain, no finetune} &  &               &  &              &  &              \\
Few-shot TPLinker (min)           &  & 2.2           &  & 3.3          &  & 2.6          \\
Few-shot TPLinker (avg)           &  & 4.5           &  & 2.9          &  & 3.5          \\
Few-shot BiTT                     &  & 7.7           &  & 1            &  & 1.7          \\ \hline
\textit{Pretrain, no finetune}    &  &               &  &              &  &              \\
Few-shot TPLinker (min)           &  & 22.8          &  & 3.2          &  & 5.6          \\
Few-shot TPLinker (avg)           &  & 21.9          &  & 3.3          &  & 5.7          \\
Few-shot BiTT                     &  & 31.9          &  & 2.0          &  & 3.8          \\ \hline
\textit{Pretrain, finetune}       &  &               &  &              &  &              \\
Few-shot TPLinker (min)           &  & 18            &  & \textbf{6.2} &  & \textbf{9.2} \\
Few-shot TPLinker (avg)           &  & \textbf{39.8} &  & 5.1          &  & 9.1          \\
Few-shot BiTT                     &  & 20.2          &  & 1.9          &  & 3.5          \\ \hline
\end{tabular}}
\caption{Main Results on the Few-NYT (INTRA) task.}\label{Tab:IntraResult}
\end{table}

The results in Table \ref{Tab:IntraResult} also support the above conclusion on Few-shot TPLinker (min). Few-shot TPLinker(min) achieves the best \textit{F1} (9.2\%) and \textit{rec} (6.2\%) scores on Few-NYT (INTRA), outperforms the second model (Few-shot TPLinker (avg)) by 0.1\% and 1.1\%. Few-shot TPLinker(avg) obtains the best \textit{Prec} (39.8\%) score, outperforms the second model (Few-shot BiTT) by 19.6\%.

\paragraph{Effect of Pre-training}
In our paper, we pretrain our encoders by appending a linear layer, and then perform softmax to achieve classification for each token. This operation can restrict the hidden state of tokens in the same class to a cluster, making the distance between these tokens small. 

As demonstrated by comparing the results of part \textit{No pretrain, no finetune} and part \textit{Pretrain, finetune} in Table \ref{Tab:InterResult} and Table \ref{Tab:IntraResult}, pre-training enables the models to learn more knowledge related to RE, which greatly boosts the performance of our models. And the improvement is reliable in different few-shot task scenarios. On Few-NYT (INTER), Few-shot TPLinker(min), Few-shot TPLinker(avg) and Few-shot BiTT boost their average \textit{F1} score respectively by 5.0\%, 14.8\% and 2.7\% after pre-training. On Few-NYT (INTRA), the three models boost their \textit{F1} separately by 6.6\%, 5.6\% and 1.8\% after pre-training. In particular, on Few-NYT (INTER), the performance boost of Few-shot TPLinker(avg) is obvious. It indicates that the few-shot training of Few-shot TPLinker(avg) has a strict requirement on the parameters initialization, while pre-training optimizes the parameters by utilizing the knowledge of the sequence tagging RE task.

\paragraph{Effect of Fine-tuning}
In our framework, the role of the loss in fine-tuneing stage is to further reduce the distance of intra-cluster tokens and increase the inter-cluster distance, thus increasing the probability of correctly classifying tokens on cluster boundaries. 

Fine-tuning on the few-shot task based on pre-training has been proved to be a appropriate method to improve the generalization of our models. Comparing the results of part \textit{ Pretrain, no finetune} and part \textit{Pretrain, finetune} in Table \ref{Tab:InterResult}, Few-shot TPLinker(min), Few-shot TPLinker(avg) and Few-shot BiTT boost their average \textit{F1} score respectively by 6.4\%, 4.6\% and 4.6\% after fine-tuning. However, experiments show that fine-tuning on the few-shot RE task may lead to overfitting in the source domain when the common knowledge between source and target domains is insufficient. For example, as shown in Table \ref{Tab:IntraResult}, on the Few-NYT (INTRA) task, the \textit{Prec}, \textit{Rec} and \textit{F1} of Few-shot BiTT separately decrease by 11.7\%, 0.1\% and 0.3\% after fine-tuning.


\paragraph{Few-NYT (INTER) vs. Few-NYT (INTRA)}
Since there is no overlap of coarse-grained relation types between the source and target domains in the Few-NYT (INTRA) task, our models is able to obtain less common information between the two domains, making it more difficult to extract triples. From Table \ref{Tab:InterResult} and Table \ref{Tab:IntraResult}, all models show a substantial decrease in performance on the Few-NYT (INTRA) task. Among them, Few-BiTT decreases the most, as evidenced by a significant drop in \textit{Prec} and a tiny \textit{Rec}. From this situation, it can be seen that the Few-BiTT model mainly learns information from the overlap of coarse-grained relation types. We conjecture that it is due to the fact that Few-BiTT needs to extract the tree structure embedded in the instance. While instances belonging to different coarse-grained types do not have similar relationship structures for Few-BiTT to learn.
\section{Conclusion}

In this paper, we put forward the definition of the few-shot RE task based on the sequence tagging approaches, and propose a few-shot RE framework for the task. Based on the few-shot RE framework, we construct three models called Few-TPLinker(min), Few-TPLinker(min) and Few-BiTT, and achieve solid results on Few-NYT (INTER) and Few-NYT (INTRA) tasks. Our future work aims to improve our frameork in two aspects, i.e. combining more category description information and exploring other ways to model the cross-domain knowledge.

\bibliography{custom}



\end{document}